\documentclass[11pt,preprint]{elsarticle}

\usepackage[cmex10]{amsmath}
\usepackage{amssymb,latexsym,xcolor,amsfonts,undertilde,cancel,mathabx}
\usepackage{graphicx,MnSymbol}
\usepackage{hyperref,wrapfig,subfig,enumerate}

\usepackage{amsthm}
\theoremstyle{plain}

\theoremstyle{definition}
\newtheorem*{definition}{Definition}
\theoremstyle{remark}

\newtheorem*{example}{Example}

\newcommand{\R}{\mathbb{R}}

\newcommand{\beq}{\begin{equation}}
\newcommand{\eeq}{\end{equation}}
\newcommand{\ba}{\begin{array}}
\newcommand{\ea}{\end{array}}
\newcommand{\bea}{\begin{eqnarray}}
\newcommand{\eea}{\end{eqnarray}}
\newcommand{\bc}{\begin{center}}
\newcommand{\ec}{\end{center}}

\newcommand{\bt}{\begin{table}}
\newcommand{\et}{\end{table}}
\newcommand{\la}[1]{\label{#1}}

\newcommand{\beqs}{\begin{equation*}}
\newcommand{\eeqs}{\end{equation*}}
\newcommand{\aln}{\begin{aligned}}
\newcommand{\ealn}{\end{aligned}}
\newcommand{\bcs}{\begin{cases}}
\newcommand{\ecs}{\end{cases}}
\newcommand{\bmat}{\begin{bmatrix}}
\newcommand{\ebmat}{\end{bmatrix}}

\newcommand{\ben}{\begin{enumerate}}
\newcommand{\een}{\end{enumerate}}

\newcommand{\jth}[1]{#1^{\text{th}}}

\definecolor{dgreen}{rgb}{0.2,0.5,0.2}
\definecolor{gold}{rgb}{0.8, 0.498, 0.196}
\definecolor{purple}{rgb}{0.502, 0, 0.502}
\definecolor{dorange}{rgb}{1,.549,0}
\definecolor{dorchid}{rgb}{0.6,.196,0.8}
\definecolor{gold}{rgb}{0.8,0.498,0.196}

\begin{document}

\begin{frontmatter}

\title{Selecting a Small Set of Optimal Gestures from an Extensive Lexicon}

\author{J.~Grosek and J.~Nathan~Kutz}

\address{Department of Applied Mathematics \\ 
University of Washington\\
Box 353925, Lewis Hall \#202 \\
Seattle, WA, 98195-3925 \\
jgrosek@uw.edu \ \ \ \ \ kutz@uw.edu}

\begin{abstract}
Finding the best set of gestures to use for a given computer recognition problem is an essential part of optimizing the recognition performance while being mindful to those who may articulate the gestures.  
An objective function, called the {\em ellipsoidal distance ratio metric} (EDRM), for determining the best gestures from a larger lexicon library is presented, along with a numerical method for incorporating subjective preferences.  
In particular, we demonstrate an efficient algorithm that chooses the best $n$ gestures from a lexicon of $m$ gestures where typically $n \ll m$ using a weighting of both subjective and objective measures.
\end{abstract}


\end{frontmatter}

\newsavebox{\SRIM}
\savebox{\SRIM}{
\begin{tabular}{|c|}
	L $\rightarrow$ 0.782 \\
	R $\rightarrow \ $ N/A \\ \hline
\end{tabular}
}



\section{Introduction}

Research in computer vision continues to be of great technological importance given the potentially vast impact in a wide range of applications in recognition software as well as human-computer interactions.  Computer vision broadly includes mathematical methods and algorithms for acquiring, processing, analyzing, and understanding images, often which are high-dimensional, in order to produce accurate decisions and classifications about what is observed~\cite{cvbook1,cvbook2}.  One increasingly important branch of the computer vision field concerns gesture recognition, where computers are trained to recognize hand signals, facial expressions, and/or eye movements in order to better interface and interact with humans~\cite{cvbook3,cvbook4}.  For many software applications that require gesture detection, classification and input,  it is often the case that only a few gestures are needed for establishing control of the software or program from among a nearly endless number of gestures that a person can articulate.  Thus given a large lexicon of gestures and an application that requires only a small subset of those gestures, one would like to know which are the best gestures to choose for the given application.  More succinctly, we develop an algorithm that chooses the best $n$ gestures from a lexicon of $m$ gestures where typically $n\ll m$.

There are many factors, both objective and subjective, which may determine which gestures are most appropriate and useful for a particular application, i.e. best gestures.  Certain gestures may be easier for the computer to process and recognize, but may not necessarily be comfortable or suitable for humans to articulate.  
Ergonomics, or the ease of articulating the various gestures, and relations to physical signs and gestures that are already in use in the culture or that are appropriate for the given application constitute {\em subjective} measures for claiming some gestures may be better than others.  The {\em objective} reasons for ranking the quality of gestures comes purely from the computer's ability to distinguish and recognize gestures in a statistical sense, i.e. from some discriminant analysis.  Both the subjective and objective reasons for determining the $n$ best gestures from a lexicon with $m$ elements need to be considered for designing gesture-based, robust software interfaces.  

As an example, consider replacing a standard computer mouse with a set of articulated gestures which use the computer's built-in camera.   The gestures would be required to replace the functions of (i) mouse movement, (ii) a one-click button, (iii) a two-click button, (iv) a scroll-up function and (v) a scroll-down function.  Thus the required lexicon set for these basic mouse functionalities gives $n = 5$.  The potential lexicon library for executing these functional behaviors is tremendously large ($m\gg 5$).  But if for the moment we limit ourselves to the sign language alphabet as potential lexicons, then $m = 26$.  Thus in this example, we would want to identify the five best sign language gestures that give the most robust statistical classification.  Such limited lexicon applications are already finding their way into the mobile phone and consumer electronics markets.  

Objectively, the best lexicons of size $n$ are determined by having the computer attempt to recognize every combination of $n$ out of $m$ gestures, and then choosing the best gesture set as the one with the highest overall successful recognition rate during a training process \cite{survey1,survey2}.  
This can be a time-consuming and combinatorially challenging problem.
In this paper a new metric, called the {\em ellipsoidal distance ratio metric} (EDRM), is introduced that provides an excellent indication as to which gestures will be easily recognizable according to the computer.  
The EDRM is applied to the feature space of the gestures, and therefore doesn't require the computer to complete the entire recognition process in order to gain some notion as to which are the best gestures in the entire lexicon of available gestures \cite{survey3}.
Thus a robust and efficient algorithm is developed to extract the best $n$ of $m$ gestures as required.

This paper is organized as follows.
First, the gesture recognition process for a computer is reviewed, and the importance of excellent feature selection is emphasized.  
Next, gesture separation in feature space is discussed as an objective measure for determining the best gestures, and the EDRM is introduced.  
A method for incorporating subjective rankings of the gestures into the EDRM is proposed.  
Then, the process of finding the best lexicons of size $n$ is demonstrated using a lexicon of static hand gestures of size $m$.  
The algorithm is evaluated in the conclusion.


\section{Gesture Representation in Feature Space}

In computer vision, the gesture recognition process starts when a raw image is imported into the computer.  
The gesture recognition process can be broken down into the following steps: 
\ben[(i)]
	\item {\em gesture detection, and segmentation}, which is the process by which the gesture is found and isolated within the image frame,
	\item {\em pre-processing}, which is the process of normalizing like-gestures to similar sizes, shapes, colors, positions, and orientations,
	\item {\em feature selection}, which is the process of determining important characteristics and aspects of gestures that will simultaneously distinguish between different gesture classes and will highlight like-gesture classes (e.g. principal component analysis), and
	\item {\em statistical learning and classification}, which is the process of training the computer to identify and recognize a gesture's articulation by statistical means and predictive functions that draw upon feature data (e.g. linear discrimination analysis).
\een
Not all recognition schemes use all of these steps, or even complete the steps in the same order as they are listed above; however, the general procedure is still valid~\cite{cvbook4}.  

 \subsection{Feature Space}

Most gesture recognition algorithms eventually represent the gestures as a sequence of features, which can be thought of as points in a feature space (See Fig.~\ref{fig:gestfs}).
This feature space provides a great deal of information about how similar or different the computer views the gestures.  
However, it is very common that most gesture recognition algorithms extract many features in order to better their recognition performance, and this corresponds to having high-dimensional feature spaces, which are usually impossible to visualize.  
The high-dimensional nature of feature spaces invokes many of the same issues that accompany the well-known {\em curse of dimensionality} \cite{curse1} found in the computer vision, statistics, and geometry fields.  
This curse is basically defined by the fact that high-dimensional data becomes intractable to systematically work with as the number of dimensions increases, especially when one does not have a preconceived notion as to how the data ought to render itself.  

\begin{figure}[t]
	\bc
   		\includegraphics[width=\linewidth]{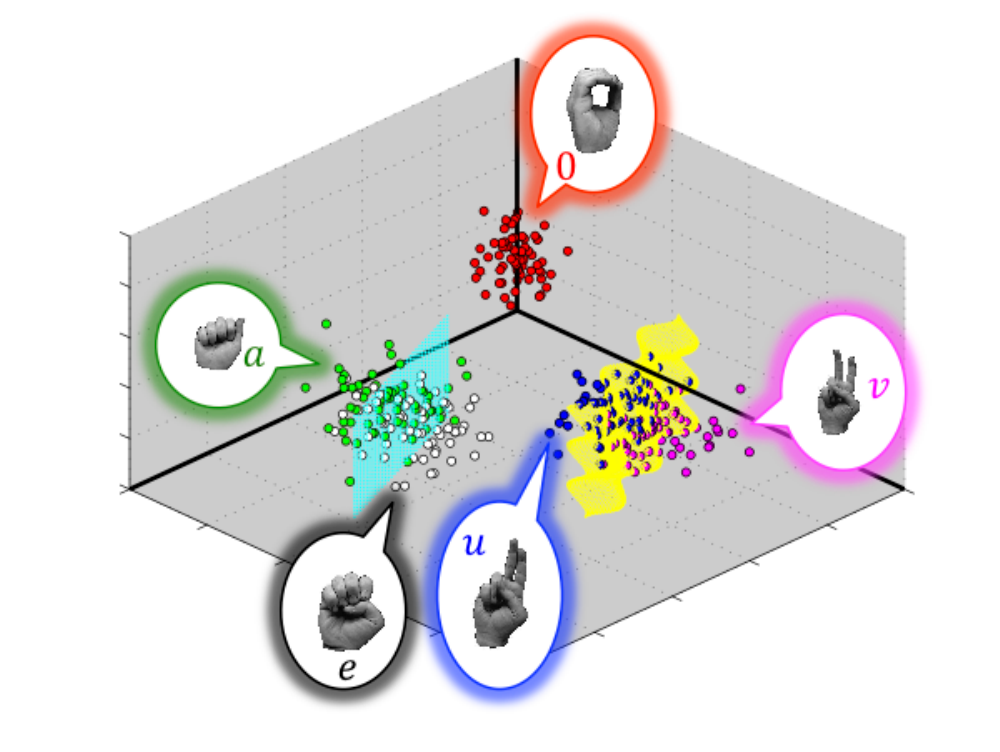}
	\ec
 	\caption{This figure exemplifies how gestures may appear in feature space.  
Gesture classes $\mathbf{a}$ (green) and $\mathbf{e}$ (white), which overlap in feature space, are separated by a linear separatrix (cyan) that attempts to distinguish the classes in some optimal sense.  
Gesture classes $\mathbf{u}$ (blue) and $\mathbf{v}$ (magenta), which do not overlap in feature space, are separated by a nonlinear separatrix (yellow).  
Gesture class $\mathbf{0}$ is well-separated from the other four classes in feature space.}
	\la{fig:gestfs}
\end{figure}

Figure~\ref{fig:gestfs} provides a visual example of what a feature space may look like; in this case, it is in 3 dimensions, which corresponds to having three features extracted from every image. 
A common way to extract features is through, for instance, a principal component analysis, but there are many other techniques available. 
Note that gesture class $\mathbf{0}$ in Fig.~\ref{fig:gestfs} is well-separated from the other four classes in feature space.  
Also, the feature points of this class are well-clustered such that there does not appear to be any images with feature points that are extreme outliers to the other images of this gesture class.  
Well-clustered feature points are a result of extracting consistent feature values from like-gestures of the same class.  
Clearly, this is a very desirable circumstance for how the feature points are positioned in feature space.  

Gesture classes $\mathbf{a}$ and $\mathbf{e}$ illustrate a common occurrence with many gestures that look similar in that these classes overlap with one another in feature space.  
Obviously, similar gestures are likely to produce similar feature values, and therefore will lie near one another in feature space.  
Also, note that classes $\mathbf{a}$ and $\mathbf{e}$ are not well-clustered, and have many outliers to their apparent centers.  
A linear separatrix, defined by $y = (2/5)x + 9/5$, can be found by the computer using statistical learning techniques in order to distinguish these two classes in some optimal sense.  
However, it is evident that there isn't a perfect way of separating these classes without likely {\em over-fitting} the data so that the separation remains general to any new data that may be used in the recognition application.  

Gesture classes $\mathbf{u}$ and $\mathbf{v}$ of Fig.~\ref{fig:gestfs} exemplify another possibility in the recognition process in that the classes are separated from one another in feature space, but the boundary between the two class regions is defined by some nonlinear function.  
In this case, a nonlinear separatrix, defined by $y = (2/5)x + \cos(4x)/4 + \cos(4z)/4 + 6$, distinguishes the two classes in some optimal sense.  
One can easily image that without enough data points, assuming they are even available, and/or without any {\em a~priori} idea as to how these classes ought to be separated, a computer algorithm may struggle to learn the nonlinear separatrix, although this may be possible with, for instance, adaptive boosting methods~\cite{boost1,boost2}.  
Even still, if the computer is able to learn data that is separated in some complex, nonlinear manner, one would still have to question whether over-fitting has occurred, and thus if the learned separation is generalizable.  
It is unlikely for most gesture recognition problems that one will be able to get any useful {\em a~priori} information as to how the features of any class will appear in feature space relative to another gesture class, especially given the high-dimensional nature of most feature spaces.  

Therefore, it would be ideal if all gesture classes were comprised of well-clustered feature points, and if they were all well-separated from one another into distinct regions of the feature space.  
One of the best ways of controlling how the gestures are rendered in feature space is by choosing excellent features in the first place.  
There is a near endless variety of features that can be extracted from a gesture, too many to make an exhaustive listing of the possibilities.   
A quality feature selection algorithm will find features that remain consistent for like-gestures and yet are very distinct for different gestures.  
In this paper, feature selection will be accomplished by either principal component analysis (PCA)~\cite{pca1, pca2, pca3} or by generalized projections (GPs)~\cite{projs1, projs2}.  

As a final note, Fig.~\ref{fig:gestfs} already highlights some of the key ideas of this work.  
In particular, if $n = 3$ gestures were required for an application from the $m = 5$ lexicon shown, then one would undoubtedly (and based upon simple intuition alone) 
choose $\mathbf{0}$ along with one of either $\mathbf{a}$ or $\mathbf{e}$ and one of either $\mathbf{u}$ or $\mathbf{v}$.  
Thus all $n = 3$ selected gestures would be well-separated for recognition and decision making in the given application.  

 \subsection{Best Features}

A common practice in the recognition field is to extract as many features as possible and allow the statistical learning algorithm to diminish the relevance of the features, usually by controlling weightings, which do not strongly effect the decision making process of the classification.  
However, in order to determine the best gestures, this practice may be detrimental because weak features are either inconsistent (noisy) for like-gestures of the same class or they are consistent across different classes.  
This inherently sabotages the chances for like-gestures to be well-clustered in feature space and yet well-separated from different gesture classes.  
Additionally, by including more features, the dimension of the feature space is increased, making it harder for both the human and computer to understand how the different gesture classes are separated in feature space.  
Thus, for the search of the best lexicons of size $n$, it is better to filter out the weak features from the analysis, and only focus on the more effective features.

Currently, there are some tests designed to evaluate the efficacy of a given feature on the entire recognition process \cite{survey3}.  
Some examples include the Fisher score~\cite{fish_score1}, the Generalized Fisher score~\cite{gen_fish_score1}, mutual information~\cite{mutual_info1}, ReliefF~\cite{relief_f1}, the Laplacian score~\cite{laplace_score1}, the Hilbert Schmidt Independence Criterion~\cite{HSIC1}, the Trace Ratio Criterion~\cite{trace_ratio1}, the Multi-Layer Perceptron Sensitivity Method~\cite{MLPSM1}, and Principle Feature Analysis~\cite{PFA1}.    

Another variable selection routine that can help determine which features have the greatest impact on the recognition performance is what termed here as the {\em feature selection weakness} (FSW). 
This measure assesses how close points of the same class/type/variety are to one another against how far points of different classes/types/varieties are separated from one another.  This is similar to the Fisher score.  
Also like the Fisher score, FSW treats the features independently and not in combination with other features.  

Assume that there are $L$ features extracted from each image and $M$ gestures in total, there being $N_{m}$ images of the $\jth{m}$ gesture, where $m \in \{ 1, 2, \ldots, M \}$.  
Therefore, the total number of images in the dataset is given by $\sum_{m = 1}^{M} N_{m}$.  
Let $X^{\ell}$ be a data structure which contains all of the feature values for feature $\ell$, where $\ell \in \{ 1, 2, \ldots, L \}$, such that $X^{\ell}$ has $M$ columns, and the $\jth{m}$ column has $N_{m}$ rows.  
The element in the $\jth{i}$ row and $\jth{m}$ column of this data structure will be denoted by $X_{im}^{\ell}$.  
The average among the feature values for feature $\ell$ and for gesture class $m$ is given by $\mu_{m}^{\ell} = \frac{1}{N_{m}}\sum_{i = 1}^{N_{m}} X_{im}^{\ell}$.  
The mean of these averages across all gesture classes is given by $\bar{\mu}^{\ell} = \frac{1}{M}\sum_{m = 1}^{M} \mu_{m}^{\ell}$.  
The variance among the features values for feature $\ell$ and for gesture class $m$ is given by $(\sigma^{2})_{m}^{\ell} = \frac{1}{N_{m} - 1}\sum_{i = 1}^{N_{m}}\left( X_{im}^{\ell} - \mu_{m}^{\ell} \right)^{2}$.  
With this notation, the FSW for the $\jth{\ell}$ feature can be defined as
\[
	\text{FSW}(\ell) = \frac{\frac{1}{M}\sum_{m = 1}^{M}\left[ (\sigma^{2})_{m}^{\ell} \right]}{\frac{1}{M - 1}\sum_{m = 1}^{M}\left( \mu_{m}^{\ell} - \bar{\mu}^{\ell} \right)^{2}}.
\]

\begin{figure*}[t]
	\bc
		\includegraphics[clip,trim=60px 30px 60px 10px,width=0.9\textwidth]{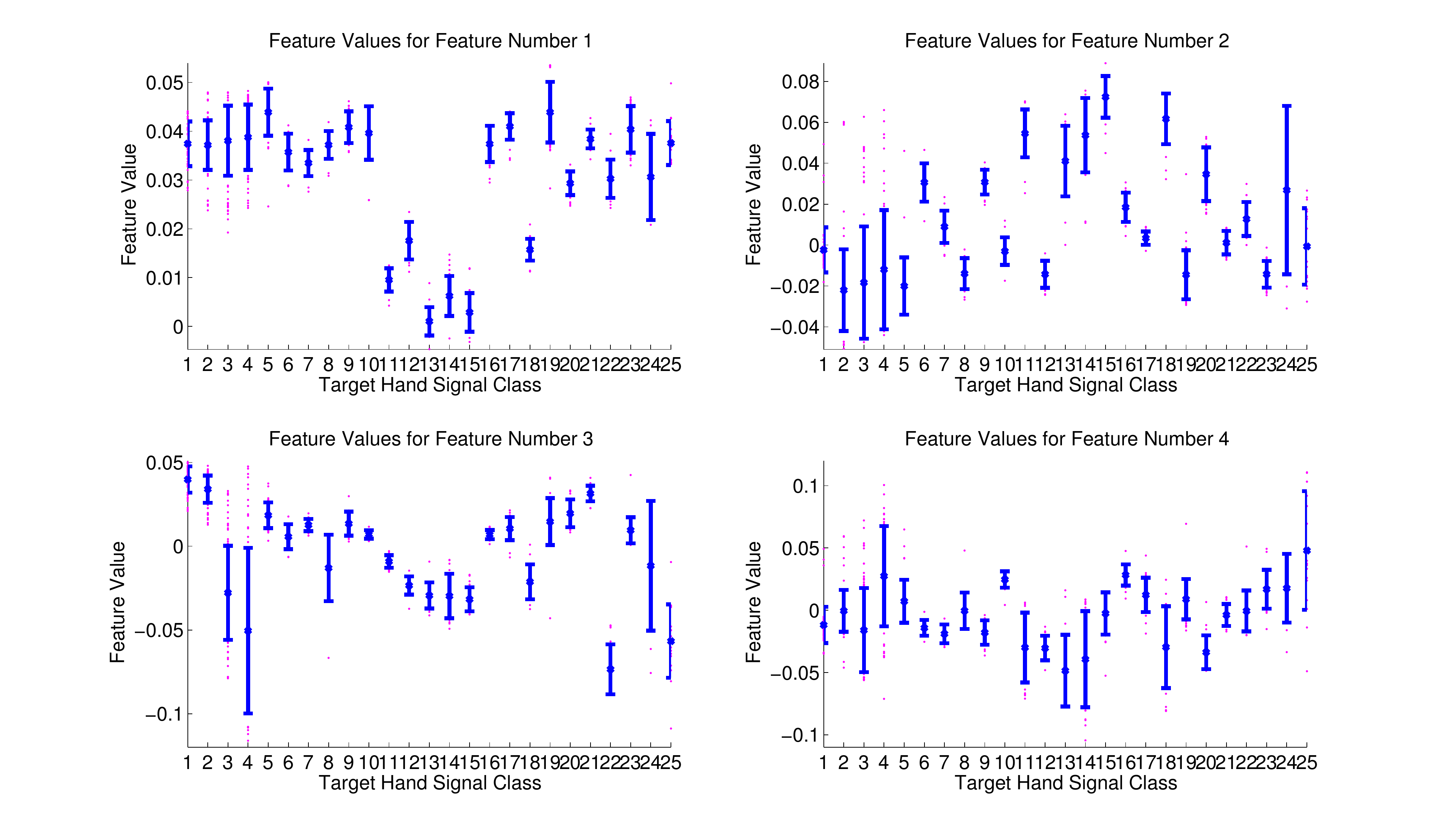}
	\ec
	\caption{Here are the feature values for the first 4 features of a recognition problem with 25 gesture classes.  
The features are extracted using the PCA method.  
The average and one standard deviation spreads (blue) of the feature values (magenta) for each class are plotted.  
Note the smaller spreads in the feature values of feature 1 as compared to feature 4, which has larger error bars.  
Also, note how, for  feature 1, some classes $\{ 11-15, 18 \}$ are well-distinguished from the rest of the classes.  
But, for feature 4, the classes are not as well-separated.  
These attributes are reflected in the FSW values shown in Fig.~\ref{fig:FSW}.}
	\la{fig:featvals}
\end{figure*}

Upon closer inspection, the numerator of the FSW value is the mean, across all classes, of the variances in each class.  
And the denominator of the FSW value is the variance, across all classes, of the means in each class.  
If the variance in the feature values within any given class is small $\left( (\sigma^{2})_{m}^{\ell} \sim \text{ small} \right)$, then it means that feature $\ell$ produces consistent values for each class, which was already noted to be a beneficial attribute for the recognition process.  
When there is large variation between the average feature values in each class $\left( \mu_{m}^{\ell} \right)$, then these classes are well-separated by feature $\ell$.  
Therefore,
\[
	\text{FSW}(\ell) = \small \frac{\text{small} = \ \stackrel{\text{in-class}}{\text{clustering}} \ = \ \stackrel{\text{in-class}}{\text{consistency}}}{\text{large} = \ \stackrel{\text{between-class}}{\text{separation}} \ = \ \stackrel{\text{between-class}}{\text{distinction}}} = (\text{small}).
\]
Figures~\ref{fig:featvals}~and~\ref{fig:FSW} illustrate how FSW values relate to the actual feature value data collected from a feature selection algorithm.  

Now it is clear that if a given feature $(\ell)$ receives a large FSW value, then this is a weak feature; hence the ``weakness'' in the term {\em feature selection weakness}.  
The contrapositive is that the better (stronger) features will  receive smaller FSW values.  
There is no definite border between ``small'' and ``large'' FSW values; rather FSW values among all the classes ought to be compared.  
The FSW value must be non-negative, and is ideally zero.  
This measure can be used to evaluate the efficacy of the features in the recognition process and even to remove the weakest features.

\begin{figure}[t]
	\bc
		\includegraphics[clip,trim=250px 30px 250px 10px,width=\linewidth]{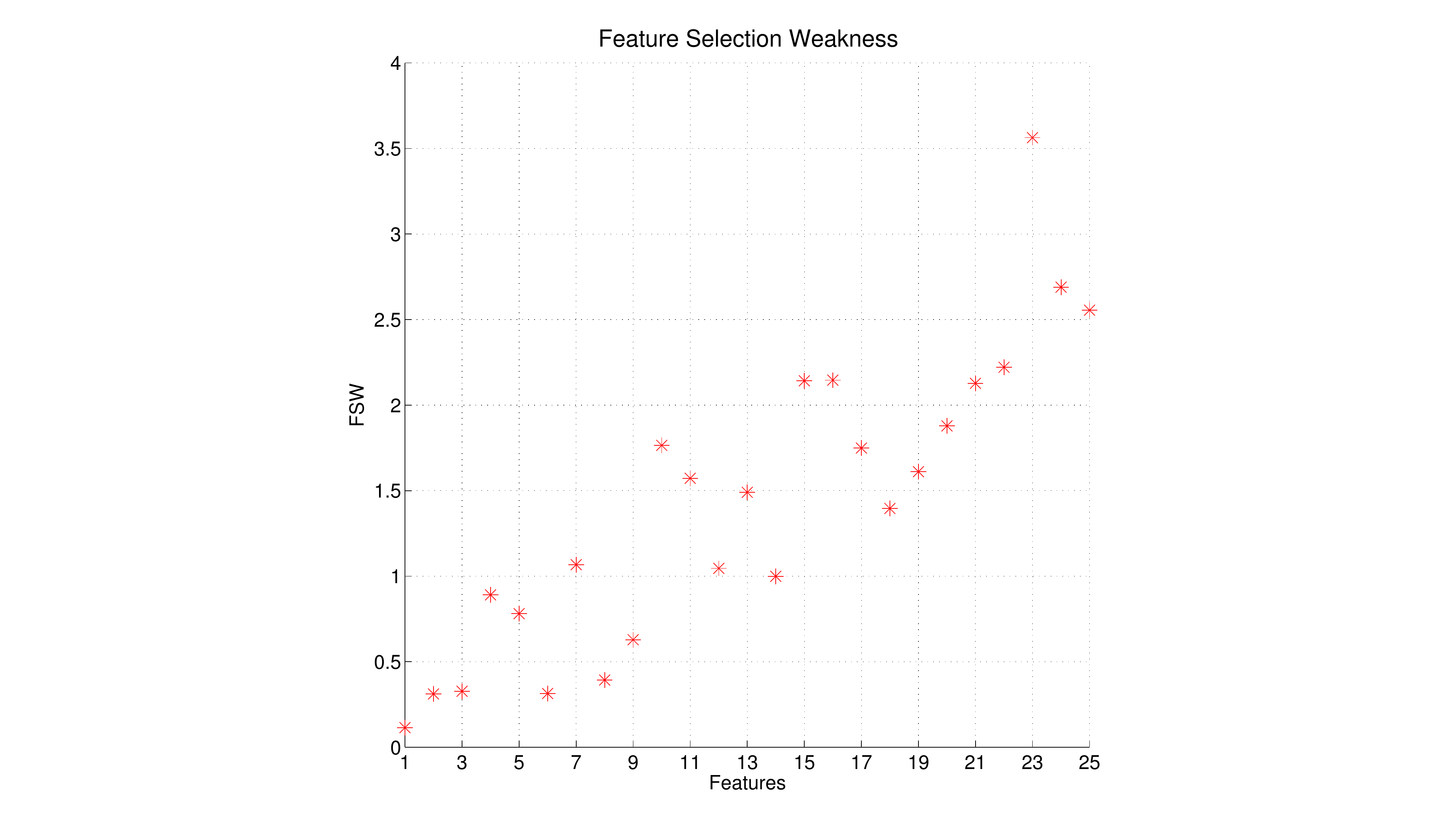}
	\ec
	\caption{Here the FSW values are plotted for the 25 features of a recognition problem that uses PCA feature selection.  
As would be expected for PCA features, the best features are extracted first, and then the features tend to get progressively weaker as more are extracted.  
The first 4 FSW values come from the feature values plotted in Fig.~\ref{fig:featvals}.}
	\la{fig:FSW}
\end{figure}


\section{Gesture Separation Measures}\la{sect:GSM}

From a computer's perspective, the best gestures are those which are the most easily distinguished through its given feature selection and statistical learning algorithms.  
For the computer, ergonomic and/or vernacular gestures are irrelevant to its consideration process.  
Previously, arguments have been made claiming that well-clustered (with-in class) and well-separated (between classes) gestures in feature space constitute easily distinguishable gestures for computer algorithms.  
However, it may be argued that with many features, even weak ones, and a sophisticated statistical learning algorithm, a computer can still achieve great recognition accuracies, even when the gestures classes are either not well-separated in feature space or are separated in some complicated, nonlinear fashion.  
There is validity to this argument, however with some caveats.  

As was alluded to previously, one major problem in the recognition field is {\em over-fitting}.  
Over-fitting occurs when a statistical learning algorithm learns its training set too well, and thus is able to distinguish between the training set gesture classes with excellent accuracies, but at the cost of not being able to maintain those accuracies in a generalized setting.  
Over-fitting can occur by simply not having a diverse enough training set, but also because of the complexity and over-sophistication of the learning algorithm.  
One might think of finding an incredibly complicated separatrix, like the one in Fig.~\ref{fig:ellipex} (B), that perfectly distinguishes two classes in the training set, but does not do as good of a job distinguishing the test data which may actually have overlapping features from the two classes.  

For most recognition applications, one cannot determine how the different classes will render themselves in feature space.  
And since many sophisticated statistical learning algorithms need some {\em a~priori} information in order for the algorithm to learn a complex, nonlinear separation measure, one is often left to either guess-work or very specific problems and applications where {\em a~priori} information is known.  

A good statistical learning algorithm may also be able to learn a nonlinear separation between classes if there are enough training data points available, especially near and well-distributed around the nonlinear separation.  
However, not every application can provide a sufficient number of training data points, nor would it be easy to guarantee that those points would be near and well-distributed around the critical areas of the feature space.  
This problem is further exacerbated by having high-dimensional data, which multiplies the number of data points needed in order to properly learn complicated class separations as the dimension increases.  
This is another reason for leaving out weak features if possible.  

In some recognition applications, especially when there are only a few gestures needed, it may be acceptable to have gesture classes that are sparse in feature space, as long as the classes are separated from one another.  
However, in general, sparse classes tend to be harder to separate than well-clustered classes \cite{survey3}.
In part, this is because most features tend to yield values in a limited range, and so sparse classes are more likely to overlap in feature space.  
Also, as more gestures are needed for the given application, it becomes more likely that a sparse gesture class will overlap in feature space with another class, making them difficult to distinguish.  

Therefore, having complicated, nonlinear separations between classes is not very desirable, nor is having sparse features in feature space.  
Thus, the idea of keeping the gestures in feature space well-clustered with-in individual classes and well-separated between different classes remains an excellent indication of how easily a computer will be able to distinguish the classes in a recognition algorithm.  
This indication remains valid and general across the many applications and specific algorithms used in the recognition process \cite{survey4}.  

 \subsection{Ellipsoidal Distance Ratio}

In order to create an objective metric that captures the desirable feature space characteristics for the recognition process of computers, it makes sense to reward gesture classes that are distributed in ellipsoidal-like shapes in feature space.  
When a gesture's features lie in an ellipsoidal distribution in feature space, they're likely to be both well-clustered and easily separated from other classes.  
The {\em ellipsoidal distance ratio metric} (EDRM), to be introduced presently, is one possible metric for gesture class separation that takes advantage of rewarding optimal class feature space distributions.  
 Here, the distinctions between the terms ellipse, ellipsoid, and hyper-ellipsoid are ignored, and the general term ellipsoid is used no matter the dimension of the object at hand.  

The first step to calculating the EDRM between two classes is to find the ellipsoid in the appropriate dimension which best encompasses the feature points for each gesture class, even, of course, if those points are not in a true ellipsoidal distribution.  
This is accomplished by finding the positive definite matrix $E$ for each class of the general ellipsoid equation: $(\vec{x} - \vec{x}_{0})^{*}E^{-1}(\vec{x} - \vec{x}_{0}) = 1$, where $(^{*})$ is the Hermitian transpose, $\vec{x}_{0}$ is the center of the ellipsoid, $\vec{x}$ defines a point on the surface of the ellipsoid, and the eigenvectors and eigenvalues of the matrix $E$ are the principal directions of the ellipsoid and the squares of the semi-axis lengths, respectively.  
The centers of the ellipsoids are of course the centers of the feature points for each class in feature space.  

Assume that the feature data for a given gesture $m$ is in a data structure $Y^{m}$ such that $Y^{m}$ has $L$ columns, which corresponds to the number of features extracted from the images, and the $\jth{\ell}$ column has $N_{m}$ rows, which corresponds to the number of images in gesture class $m$.  
Note that the number of features extracted $L$ is also the dimension of the feature space.  
By the properties of the {\em singular value decomposition} (SVD), the ellipsoidal matrix for gesture $m$ is $E_{m} = \left( Y^{m} \right)^{*} \cdot Y^{m}$.  
This works out because the SVD states that any matrix $Z \in \R^{q \times r}$ can be decomposed such that $Z = U\Sigma V^{*}$, where $U \in \R^{q \times q}$ is unitary, $\Sigma \in \R^{q \times r}$ is diagonal, and $V \in \R^{r \times r}$ is unitary.  
This implies that $Z \cdot Z^{*} = U\Sigma^{2}U^{*}$, which is the eigendecomposition of $Z \cdot Z^{*}$.  
So if the ellipsoid matrix $E = Z \cdot Z^{*}$, then the principal directions of the ellipsoid are found in the columns of $U$ and the corresponding squares of the semi-axis lengths are along the main diagonal of $\Sigma^{2}$.  
Letting $Z = \left( Y^{m} \right)^{*}$, the ellipsoidal matrix $E$ is guaranteed to have the proper dimensions $(L \times L)$ and proper attributes that best fit the feature data of feature $m$ into an ellipsoid.  

Obviously, the more gesture images in each class, the better sampling one will get for the true size and nature of the class in feature space.  
It would be good to have at least as many images in each gesture class as there are dimensions of the feature space $(L)$ in order to create the gesture ellipsoids and to calculate the EDRM.  

At this point one has the option of rescaling the size of the ellipsoid for each gesture class.  
This somewhat arbitrary rescaling can have significant effects on the EDRM.  
Clearly, it is best to choose a consistent rescaling scheme across all the gesture classes in order to maintain fair comparisons of the class sizes and distances in feature space.  
It is suggested that each ellipsoid's surface be a certain percentage of the distance from the center of the ellipsoid to the most extreme outlier feature point in that gesture class.  
In this way, it is guaranteed that at least some feature points from each class are on the outside of the ellipsoid of its own class.  
This method acknowledges the reality that there are almost always impure articulations (outliers) of any gesture that may not fully represent the class.  
The actual percentage used in this convention can be determined based on how pure the gesture images are for each class.  
For instance, if all the images used in the dataset are of well-articulated gestures, then one might choose to have the ellipsoid encompass most of the feature points; this is assuming there are few to no outliers, and so a high percentage value is used, e.g. $90\%-100\%$.  
Whereas, a noisier sampling of the gestures might assume the existence of outliers in each class, and so a lower percentage of $60\%-80\%$ is used.  
In what follows, the convention will be set at $65\%$.  

The {\em ellipsoidal distance ratio} is the shortest distance between two ellipsoids compared against the distance between the ellipsoid centers.  
Measuring the straight-line distance between ellipsoid centers is trivial.  
However, the shortest distance between two ellipsoids needs to be formulated as an optimization problem.  
Suppose there are two classes, $\mathbf{a}$ and $\mathbf{b}$, with ellipsoid surface points $\vec{x}^{\mathbf{a}}$ and $\vec{x}^{\mathbf{b}}$, centers $\vec{x}_{0}^{\mathbf{a}}$ and $\vec{x}_{0}^{\mathbf{b}}$, and ellipsoid matrices $E_{\mathbf{a}}$ and $E_{\mathbf{b}}$, respectively.  
The optimization problem can then be stated as follows
\begin{align*}
	\arg \min & \sqrt{\left( \vec{x}^{\mathbf{a}} - \vec{x}^{\mathbf{b}} \right)^{*}\cdot \left( \vec{x}^{\mathbf{a}} - \vec{x}^{\mathbf{b}} \right)} \\
	\text{subject to } & \left( \vec{x}^{\mathbf{a}} - \vec{x}_{0}^{\mathbf{a}} \right)^{*}E_{\mathbf{a}}^{-1}\left( \vec{x}^{\mathbf{a}} - \vec{x}_{0}^{\mathbf{a}} \right) = 1 \ \& \\
	& \left( \vec{x}^{\mathbf{b}} - \vec{x}_{0}^{\mathbf{b}} \right)^{*}E_{\mathbf{b}}^{-1}\left( \vec{x}^{\mathbf{b}} - \vec{x}_{0}^{\mathbf{b}} \right) = 1.
\end{align*}
This problem effectively says that one wants the shortest distance between the points $\vec{x}^{\mathbf{a}}$ and $\vec{x}^{\mathbf{b}}$, while constraining $\vec{x}^{\mathbf{a}}$ to be on the surface of the ellipsoid of gesture $\mathbf{a}$ and constraining $\vec{x}^{\mathbf{b}}$ to be on the surface of the ellipsoid of gesture $\mathbf{b}$.  
Without any loss of purpose or accuracy, the optimization problem can be better formulated for numerical and algorithmic stability issues such that
\begin{align*}
	\arg \min & \frac{\left( \vec{x}^{\mathbf{a}} - \vec{x}^{\mathbf{b}} \right)^{*}\cdot \left( \vec{x}^{\mathbf{a}} - \vec{x}^{\mathbf{b}} \right)}{2} \\
	\text{subject to } & \left[ \left( \vec{x}^{\mathbf{a}} - \vec{x}_{0}^{\mathbf{a}} \right)^{*}E_{\mathbf{a}}^{-1}\left( \vec{x}^{\mathbf{a}} - \vec{x}_{0}^{\mathbf{a}} \right) - 1 \right]^{2} = 0 \ \& \\
	& \left[ \left( \vec{x}^{\mathbf{b}} - \vec{x}_{0}^{\mathbf{b}} \right)^{*}E_{\mathbf{b}}^{-1}\left( \vec{x}^{\mathbf{b}} - \vec{x}_{0}^{\mathbf{b}} \right) - 1 \right]^{2} = 0.
\end{align*}

Since, in high-dimensions, these ellipsoids cannot be easily visualized, one needs to take extra precautions to ensure that the ellipsoids do not overlap in feature space.  
Given the optimal points $\tilde{\vec{x}}^{\mathbf{a}}$ and $\tilde{\vec{x}}^{\mathbf{b}}$, one check that can be made is to ensure that
\begin{align*}
	\left( \tilde{\vec{x}}^{\mathbf{b}} - \vec{x}_{0}^{\mathbf{a}} \right)^{*}E_{\mathbf{a}}^{-1}\left( \tilde{\vec{x}}^{\mathbf{b}} - \vec{x}_{0}^{\mathbf{a}} \right) & > 1 \ \& \\
	\left( \tilde{\vec{x}}^{\mathbf{a}} - \vec{x}_{0}^{\mathbf{b}} \right)^{*}E_{\mathbf{b}}^{-1}\left( \tilde{\vec{x}}^{\mathbf{a}} - \vec{x}_{0}^{\mathbf{b}} \right) & > 1.
\end{align*}
These inequalities verify that the optimal point on ellipsoid $\mathbf{b}$ is not inside of the ellipsoid for gesture $\mathbf{a}$, and vice-versa.  
Also, due to numerical inaccuracies it can be good to verify that
\begin{align*}
	\Bigg| 1 - \left( \tilde{\vec{x}}^{\mathbf{a}} - \vec{x}_{0}^{\mathbf{a}} \right)^{*}E_{\mathbf{a}}^{-1}\left( \tilde{\vec{x}}^{\mathbf{a}} - \vec{x}_{0}^{\mathbf{a}} \right) \Bigg| & < \epsilon \ \& \\
	\Bigg| 1 - \left( \tilde{\vec{x}}^{\mathbf{b}} - \vec{x}_{0}^{\mathbf{b}} \right)^{*}E_{\mathbf{b}}^{-1}\left( \tilde{\vec{x}}^{\mathbf{b}} - \vec{x}_{0}^{\mathbf{b}} \right) \Bigg| & < \epsilon,
\end{align*}
where $\epsilon$ is some small tolerance for numerical inaccuracy, which may need to grow with the number of dimensions of the problem.  
These inequalities ensure that the optimal point for ellipsoid $\mathbf{a}$ is truly on the surface of ellipsoid $\mathbf{a}$, and the same for ellipsoid $\mathbf{b}$.  

Once these points on the two gesture class ellipsoids that minimize the distance between the ellipsoids have been found, one can calculate the {\em ellipsoidal distance ratio metric} as follows
\[
	\text{EDRM}_{\mathbf{a}, \mathbf{b}} = \frac{\sqrt{\left( \tilde{\vec{x}}^{\mathbf{a}} - \tilde{\vec{x}}^{\mathbf{b}} \right)^{*}\cdot \left( \tilde{\vec{x}}^{\mathbf{a}} - \tilde{\vec{x}}^{\mathbf{b}} \right)}}{\sqrt{\left( \vec{x}_{0}^{\mathbf{a}} - \vec{x}_{0}^{\mathbf{b}} \right)^{*}\cdot \left( \vec{x}_{0}^{\mathbf{a}} - \vec{x}_{0}^{\mathbf{b}} \right)}}.
\]
The EDRM values are always between 0 and 1.  
Larger EDRM values indicate that the two classes that are being compared are further separated and better clustered in feature space.  
Also note that the EDRM normalizes the scale of the ellipsoids in that larger ellipsoids must be more separated from one another in order to have the same EDRM score as smaller ellipsoids.  
If the volumes of both ellipsoids go to zero, because of the clustering of feature points, then the EDRM approaches its ideal value of 1 because the shortest distance between the ellipsoid surfaces (numerator) becomes the same as the distance between the ellipsoid centers (denominator).  
 
\begin{figure}[t]
	\bc
   		\includegraphics[width=\linewidth]{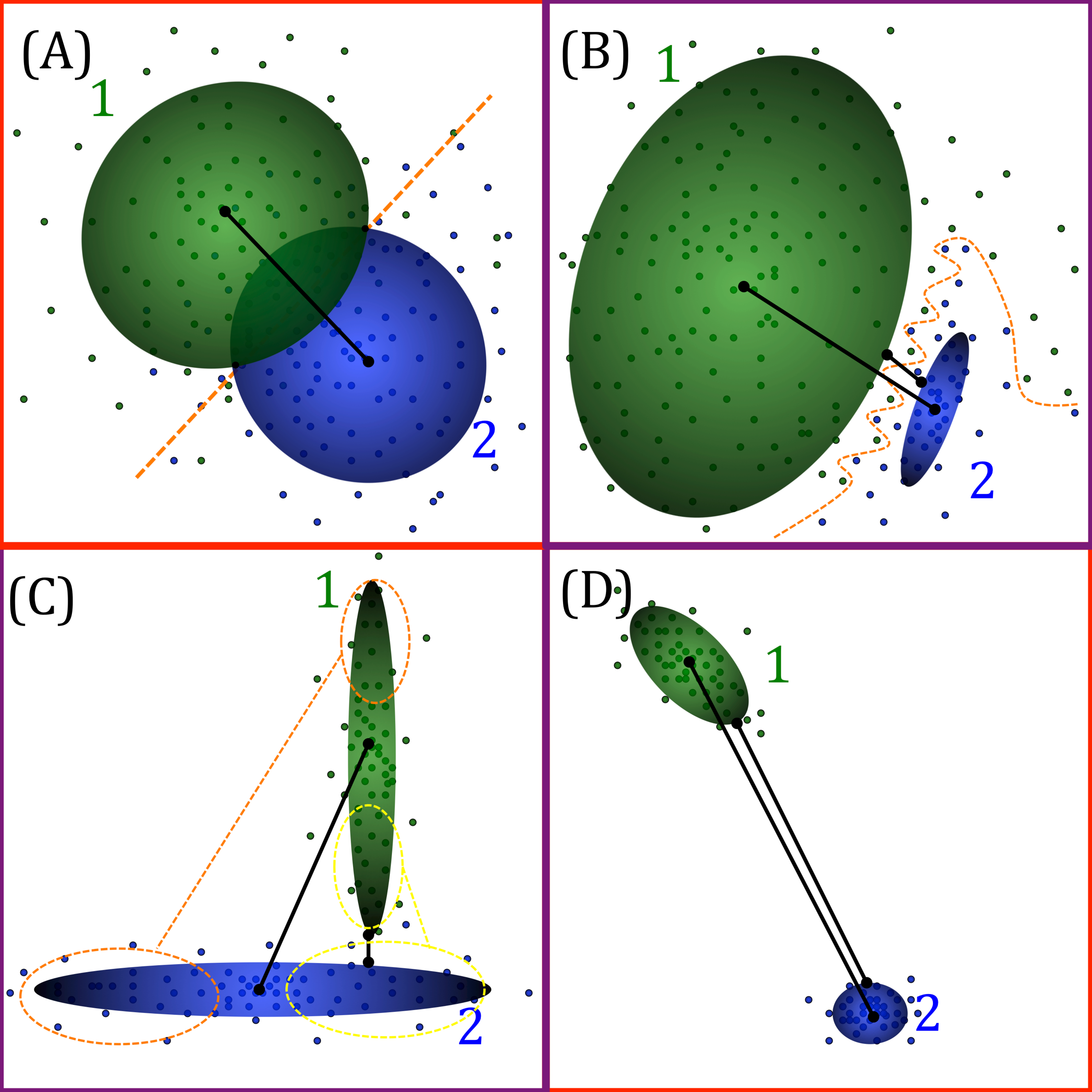}
	\ec
 	\caption{In this figure four examples illustrate why ellipsoidal distance ratios (EDRs) are good measures of class separation in feature space.  
The EDR is the shortest distance between two ellipsoids divided by the distance between the ellipsoid centers (black lines).  
When two classes overlap (A), so do their corresponding ellipsoids; 
thus, the EDR is zero.  
Even when two classes have non-ellipsoid distributions of feature points and are separated by a nonlinear separatrix (B), the EDR still provides a reliable measure metric for closeness.  
The EDR measure isn't easily fooled when many points from the two classes are well separated, but some are not (C) - 
compare the distances between features points in the upper half of class $\mathbf{1}$ and in the left half of class $\mathbf{2}$ to the distances using the lower half of class $\mathbf{1}$ and the right half of class $\mathbf{2}$.  
Ideally, each class is well-clustered and well-isolated from the other class (D), thus yielding a EDR value near one.}
	\la{fig:ellipex}
\end{figure}
 
Figure~\ref{fig:ellipex} illustrates why the EDRM is a good metric for gesture separation.  
First note that when feature points from the two given classes significantly overlap (A), the EDR value will always be zero because the class ellipsoids will also overlap and so the distance between the ellipsoid surfaces is zero.  
In the case that one or both feature point regions for the two classes have non-ellipsoidal distributions (B), the best-fitting ellipsoids to the class regions are determined.  
Further, in the case that these regions are still fairly well-separated, a EDRM score can be calculated that indicates how well-separated the regions are.  
In some instances there can be sub-regions of the two classes that are well-separated in feature space, and yet other sub-regions of the two classes in feature space that are nearby each other (C).  
In this scenario, the EDRM will produce a practical value that accounts for the different sub-regions and their respective separations.  
The EDRM is neither misled into being too high nor too low by the positioning of the sub-regions.  
Ideally, the two classes are well-clustered and well-separated (D).

 \subsection{Subjective Separation Measures}\la{sect:SSM}

For many applications, the comfort and ease of articulating a gesture can be an important factor for determining which gestures are more appropriate for use.  
Moreover, some gestures better suite an application because of the cultural norms and meanings of the gesture already in use in society.  
In order to incorporate these ergonomic and vernacular reasons into the best gestures decision making process, one would need to subjectively rate all the gestures in the entire lexicon of available gestures.  

One method of creating a subjective measure for ranking the suitability of the gestures that is comparable with the EDRM, is by first rating the ergonomic and vernacular quality of each gesture on a scale of $[0, 1]$.  
A rating of $0$ would mean {\em not suitable} for the given application, and a rating of $1$ would mean {\em most suitable} for the task.  
The fact that this rating scale is the same as the range of possible EDRM scores is not coincidental.  

Note that the EDRM scores are comparisons between two classes, therefore the subjective ratings must also be a comparisons between two classes in order to be able to integrate this subjective measure with the EDRM.  
This is accomplished by creating a {\em subjective measure} (SM) for every class pairing that is the average of the numerical ratings that each class in the pair was given.  
\[
	\text{SM}_{\mathbf{a}, \mathbf{b}} = \frac{\text{Rating}(\mathbf{a}) + \text{Rating}(\mathbf{b})}{2} \in [0,1]
\]
Table~\ref{tab:subjex} is a four class lexicon example of how subjective measures are calculated.  Note that we subjectively 
generated the values of this table.  Ultimately, one might refine this process by having a large sample of people vote on their favorite gestures, thus generating average scores of gesture likability across a population.  

\bt[b]
	\bc
		\caption{Here a lexicon of 4 gestures, labeled $\mathbf{a}-\mathbf{d}$, receives subjective ratings, written in parentheses, which are used to create subjective measures (SM) between each class pairing.}
		\la{tab:subjex}
		\begin{tabular}{|c|c|c|c|c|}
			\hline
			$\stackrel{\mbox{}}{\mbox{}}$ & $\stackrel{\mathbf{a}}{(0.9)}$ & $\stackrel{\mathbf{b}}{(0.2)}$ & $\stackrel{\mathbf{c}}{(0.7)}$ & $\stackrel{\mathbf{d}}{(0.5)}$ \\ \hline
			$\stackrel{\mathbf{a}}{(0.9)}$ & - & 0.55 & 0.80 & 0.70 \\ \hline
			$\stackrel{\mathbf{b}}{(0.2)}$ & 0.55 & - & 0.45 & 0.35 \\ \hline
			$\stackrel{\mathbf{c}}{(0.7)}$ & 0.80 & 0.45 & - & 0.60 \\ \hline
			$\stackrel{\mathbf{d}}{(0.5)}$ & 0.70 & 0.35 & 0.60 & - \\ \hline
		\end{tabular}
	\ec
\et

The subjective measure and the EDRM for each class pairing can be combined into a {\em total measure} (TM) by a weighting factor $\alpha \in [0,1]$, which determines the relative influence that the subjective measure will have on the total measure.  
\[
	\text{TM}_{\mathbf{a}, \mathbf{b}} = \alpha\text{SM}_{\mathbf{a}, \mathbf{b}} + (1 - \alpha)\text{EDR}_{\mathbf{a}, \mathbf{b}} \in [0,1]
\]

If any gesture must be included in the best lexicon of size $n$, then the problem becomes a matter of finding  the best lexicon of size $n - 1$, where the $\jth{n} $ gesture is already determined.  
In the case that one of $p$ $(p \ll n)$ gestures must be included, in the best lexicon of size $n$, then this constraint can be enforced in the programming of the algorithmic search process.  
Finally, if there is a gesture or multiple gestures that should never be chosen to be a part of the best gestures set, then these gestures can simply be removed from the entire lexicon of available gestures, and therefore from the process of consideration.  

 \subsection{Best Gestures}

Once one has calculated the total measure (TM) for every gesture class pairing in the entire lexicon of available gestures, determining the best gesture subset is a fairly straight-forward process.  
Assume there are a total of $m$ available gestures in our lexicon library, and one desires to know what is the best lexicon of size $n$, where $n < m$.  
The best lexicon of size 2 is simply the gesture class pairing with the largest TM value.
In order to calculate the best lexicon of size $n \geq 3$, one must first find all the unique combinations of $n$ gestures in the set of $m$ possible gestures.  
Then, the sum of the TMs from each pairing in every unique combination of $n$ gestures is calculated.  
The unique grouping of $n$ gestures with the highest total sum TM value constitutes the best lexicon of size $n$.  

Note that this problem is {\em NP hard} because there are 
\newsavebox{\mchoosen}
\savebox{\mchoosen}{$
\begin{pmatrix}
	m \\
	n
\end{pmatrix}
$}
\[
	\usebox{\mchoosen} = \frac{m!}{n!(m - n)!}
\]
unique combinations of size $n$, and 
\newsavebox{\nchooset}
\savebox{\nchooset}{$
\begin{pmatrix}
	n \\
	2
\end{pmatrix}
$}
\[
	\usebox{\nchooset} = \frac{n!}{2!(n - 2)!}
\]
pairings within each combination of $n$.  
Therefore, determining the best gesture set of size $n \geq 3$ requires at least
\[
	\usebox{\mchoosen} \cdot \usebox{\nchooset} = \frac{m!}{2!(n - 2)!(m - n)!}
\]
operations, which grows unsustainably with $m$, and the problem becomes intractable.  
However, the process is still manageable when $n$ is small relative to $m$.  

Note that the subjective metric (SM) is set up in such a way that the best lexicon of size $n$ will always include the best lexicon of size $n - 1$; thus, one only needs to find the best gesture to add to the $n - 1$ already found.  
This makes the problem much more tractable.  
However, the objective EDRM does allow for situations where the best lexicon of size $n$ does not include the best lexicon of size $n - 1$.  
This can occur for a variety of reasons including cases where there are two similar gestures, which of course are nearby in feature space, and one of them is better separated from a few other gestures.  
But, as the number of gestures included in the best gesture set increases, the other gesture of the two like-gestures becomes the better choice.  
Another example is when two different gestures are very well-separated from one another in feature space; however, one of the two gestures is fairly close to the other available gestures in the entire lexicon.  
In this case, the best lexicon of 2 is determined by the two aforementioned well-separated gestures, but the best lexicons of $n \geq 3$ will not include the gesture that is close to the rest of the available gestures.  
In any case, assuming that the best lexicon of size $n$ includes the best lexicon of size $n - 1$ simplifies the search process, but at the cost of potentially not finding the true best lexicon of size $n$.  


\section{Example and Results}

In order to test this process for determining the best gesture sets, consider a real static hand gesture recognition problem.  
For this example, consider the Massey University ASL static hand gesture dataset~\cite{dataset1, dataset2} of 36 hand gestures, which has 70 renditions of each gesture class, except for class $\mathbf{t}$ which has 65 renditions (See Fig.~\ref{fig:allgests}).  
The hands have already been isolated within each image and segmented with black backgrounds.  
The raw images are pre-processed by converting to grayscale, centering the hand within the frame, down-sampling the images to $32 \times 32$ (without changing the aspect ratio of the hand), and normalizing the brightest pixel intensity~\cite{prep1}.  
As has been stated, feature selection is done by the PCA method and the GP method, both only using 10 features.  
Statistical learning will be done by the LDA method~\cite{LDA1}, with a one-vs-the-rest classification style~\cite{prep1}.  
The training set consists of 20 randomly chosen images from each gesture class, and the test data consists of the remaining 50 images (45 for class $\mathbf{t}$) from each class.  

\begin{figure}[t]
	\bc
   		\includegraphics[width=\linewidth]{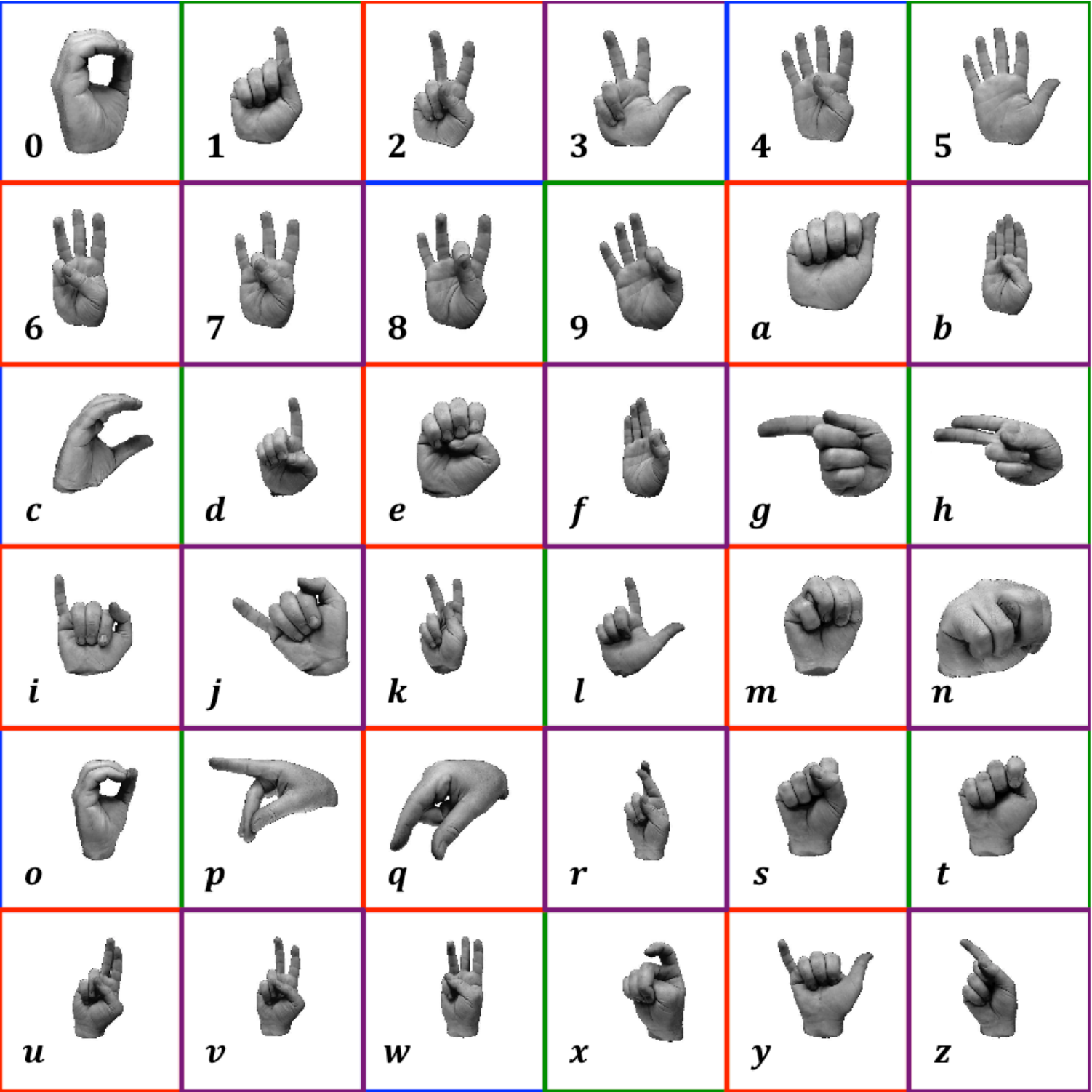}
	\ec
 	\caption{The ASL static hand gesture dataset from Massey University~\cite{dataset1, dataset2} is a lexicon of 36 gesture classes, labeled $\mathbf{0-9}$, then $\mathbf{a-z}$.}
	\la{fig:allgests}
\end{figure}

For the case where there are no subjective considerations to be incorporated into the decision making process $(\alpha = 0)$, the computer will rely solely on the objective EDRM to determine the best lexicon of size $n$.  
Figure~\ref{fig:best0} illustrates the results of this experiment using both the PCA (left) and GP (right) feature selection methods.  
Because these methods produce profoundly different features, the best gesture sets that are determined are also different.  

Of course, the true best gesture sets, without subjective considerations, are those which produce the highest recognition rates from the given algorithms.  
Recognition accuracy can take on two forms: (1) the {\em within class success rate}, meaning when images are correctly labeled within their respective class, and (2) the \textit{out-of-class success rate}, meaning when images are correctly not labeled to belong to classes to which they do not belong.  
By counting the number of times each image is correctly labeled within its own class and not labeled to belong to another class, the {\em within class} and \textit{out-of-class} success rates can be calculated and then averaged together to create an overall average recognition rate.  
By finding this overall average recognition rate for every possible combination of $n$ gestures, and picking the maximal rate, one thus discovers the corresponding true best gesture set of size $n$.  

Using this prescribed method, all of the best lexicons of size $n$ in Fig.~\ref{fig:best0} that were determined by the EDRM have been confirmed to also be the true best lexicons.  
For all of the different cases presented in Fig.~\ref{fig:best0}, there are multiple lexicons of size $n$ that tied for being the best gesture set according to the average recognition rates (only one is shown), which isn't surprising for small lexicons.  

One caveat to mention is that the LDA statistical learning method relies on Fisher linear discriminants, which attempt to group like-classes while separating different classes in a projection from feature space onto a line.  
Since the EDRM rewards pairs of classes that are both well-clustered and well-separated in feature space, the LDA method is more likely to produce good recognition rates on the same classes that bode well under the EDRM.  
Other statistical learning algorithms may produce slightly different best lexicons, but are still very likely to rate the best lexicons found using the EDRM very highly, for the reasons mentioned earlier in the beginning of Section~\ref{sect:GSM}.  

\begin{figure}[t]
	\bc
   		\includegraphics[width=\linewidth]{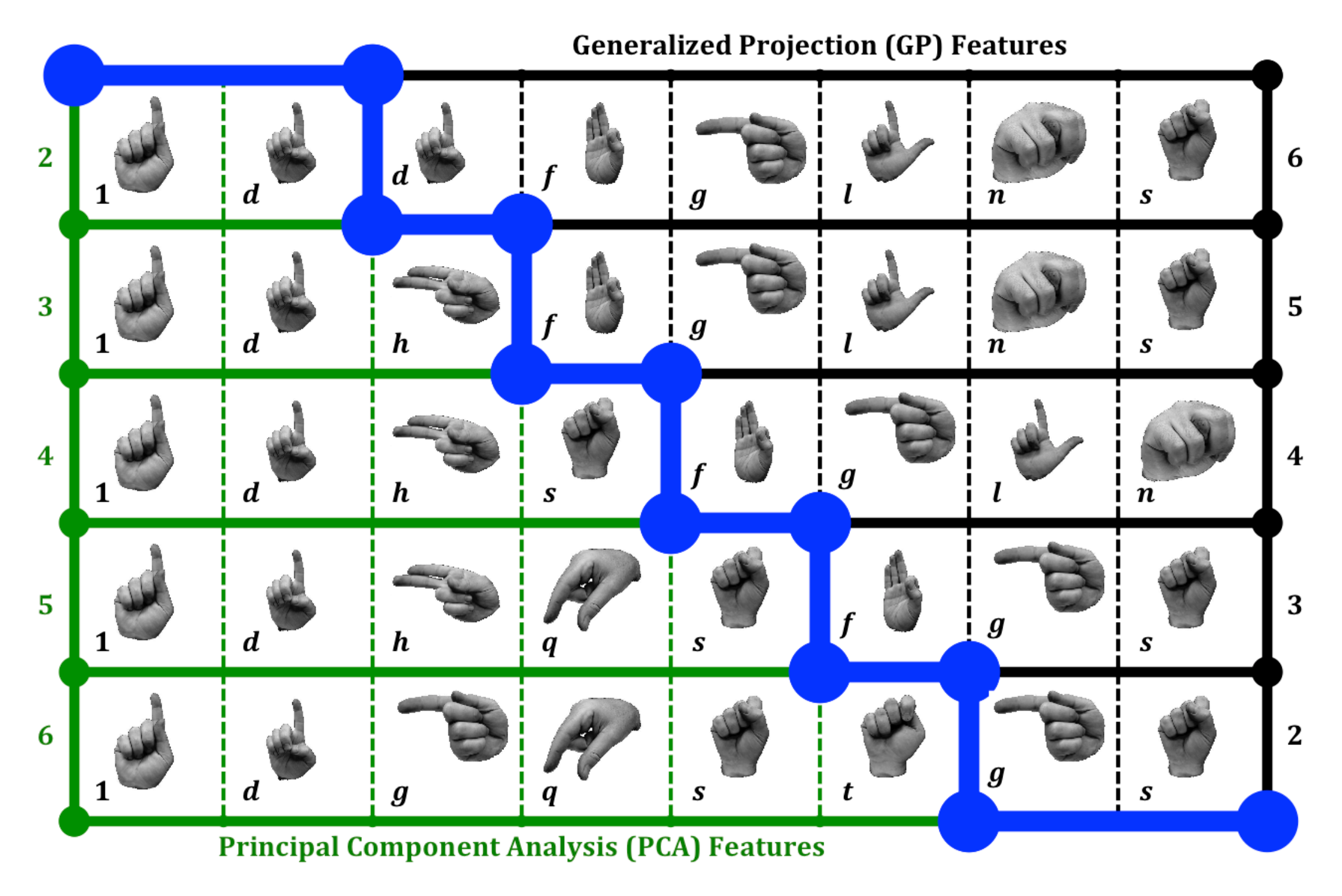}
	\ec
 	\caption{These are the best lexicons of sizes $n=2-6$ using the PCA (green) and GP (black) feature extraction methods.  
The lexicons of size $n$ are read across each row; the blue line separates the two feature selection methods.  
In this scenario $\alpha = 0$, which means that any subjective considerations for determining better gestures are being ignored and the total measure (TM) comes solely for the EDRM.}
	\la{fig:best0}
\end{figure}

One can incorporate subjective considerations into the TM metric as was described in Section~\ref{sect:SSM}.  
Table~\ref{tab:subjranks} lists an example of possible subjective rankings given to each gesture for the purposes of this experiment.  
Using these rankings, the SM for each class pairing is calculated and incorporated into the TM according to the weighting $\alpha > 0$.  
Figure~\ref{fig:bestsubj} displays the best lexicons of sizes $2-6$ for the four scenarios: (a)~$\alpha = 0.25$, (b)~$\alpha = 0.50$, (c)~$\alpha = 0.75$, and (d)~$\alpha = 1$.  

\bt[b]
	\bc
		\caption{The subjective rankings given to each gesture in the Massey University lexicon are listed, from which the subjective measures are calculated.}\la{tab:subjranks}
		\scalebox{0.85}{
		\begin{tabular}{|c|c|c|c|c|c|}
			\hline
			$\mathbf{0} \rightarrow 0.97$ & $\mathbf{1} \rightarrow 0.10$ & $\mathbf{2} \rightarrow 1.00$ & $\mathbf{3} \rightarrow 0.30$ & $\mathbf{4} \rightarrow 0.95$ & $\mathbf{5} \rightarrow 0.96$ \\ \hline
			$\mathbf{6} \rightarrow 0.90$ & $\mathbf{7} \rightarrow 0.45$ & $\mathbf{8} \rightarrow 0.55$ & $\mathbf{9} \rightarrow 0.50$ & $\mathbf{a} \rightarrow 0.85$ & $\mathbf{b} \rightarrow 0.85$ \\ \hline
			$\mathbf{c} \rightarrow 0.95$ & $\mathbf{d} \rightarrow 0.99$ & $\mathbf{e} \rightarrow 0.85$ & $\mathbf{f} \rightarrow 0.60$ & $\mathbf{g} \rightarrow 0.90$ & $\mathbf{h} \rightarrow 0.95$ \\ \hline
			$\mathbf{i} \rightarrow 0.70$ & $\mathbf{j} \rightarrow 0.20$ & $\mathbf{k} \rightarrow 0.10$ & $\mathbf{l} \rightarrow 0.95$ & $\mathbf{m} \rightarrow 0.90$ & $\mathbf{n} \rightarrow 0.85$ \\ \hline
			$\mathbf{o} \rightarrow 0.97$ & $\mathbf{p} \rightarrow 0.75$ & $\mathbf{q} \rightarrow 0.70$ & $\mathbf{r} \rightarrow 0.45$ & $\mathbf{s} \rightarrow 0.95$ & $\mathbf{t} \rightarrow 0.90$ \\ \hline
			$\mathbf{u} \rightarrow 0.95$ & $\mathbf{v} \rightarrow 0.98$ & $\mathbf{w} \rightarrow 0.90$ & $\mathbf{x} \rightarrow 0.55$ & $\mathbf{y} \rightarrow 0.30$ & $\mathbf{z} \rightarrow 0.30$ \\ \hline
		\end{tabular}
		}
	\ec
\et

\begin{figure*}[t]
	\bc
		\subfloat[][$\alpha = 0.25$]{\la{fig:best25}\includegraphics[width=0.45\linewidth]{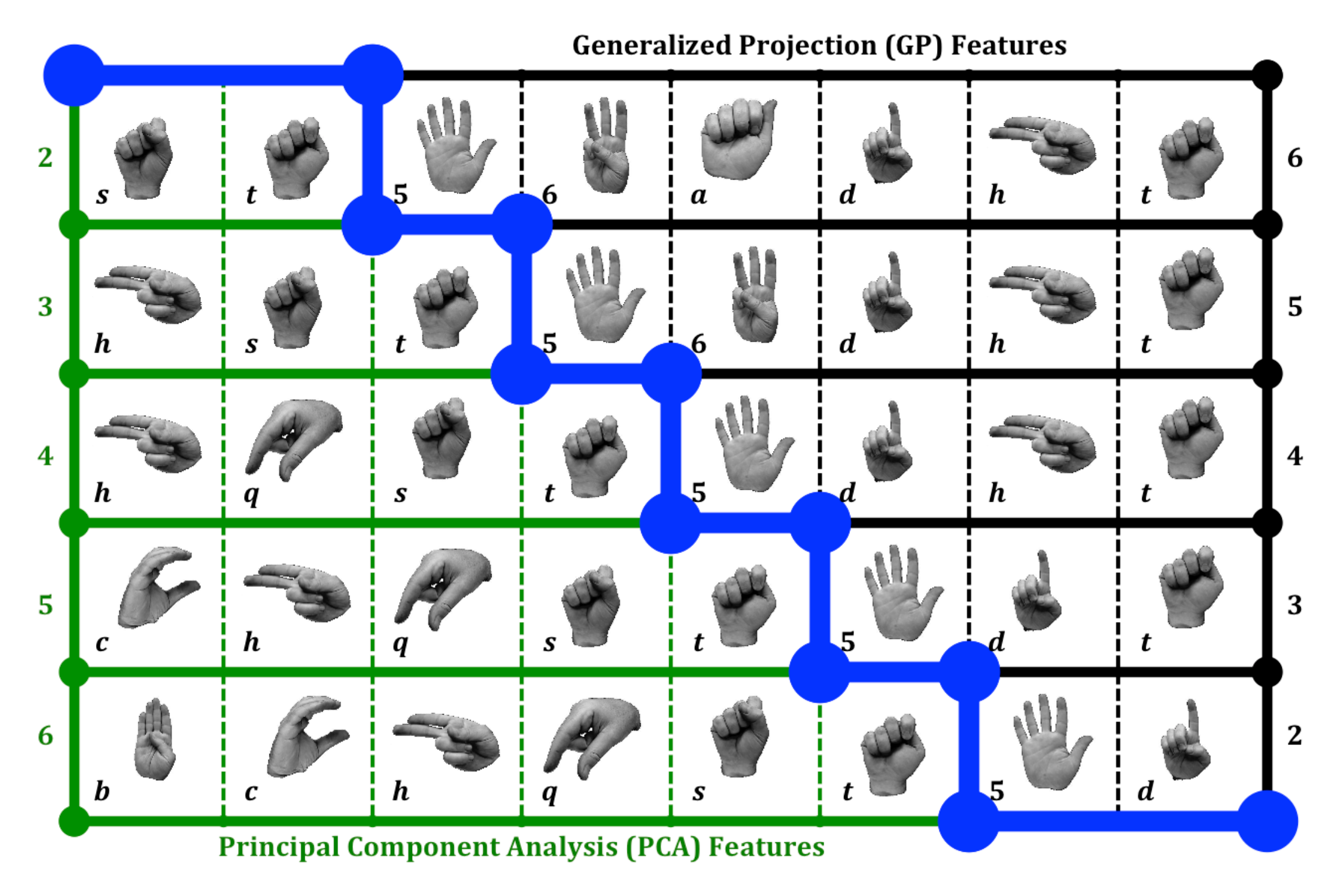}} \quad
		\subfloat[][$\alpha = 0.50$]{\la{fig:best50}\includegraphics[width=0.45\linewidth]{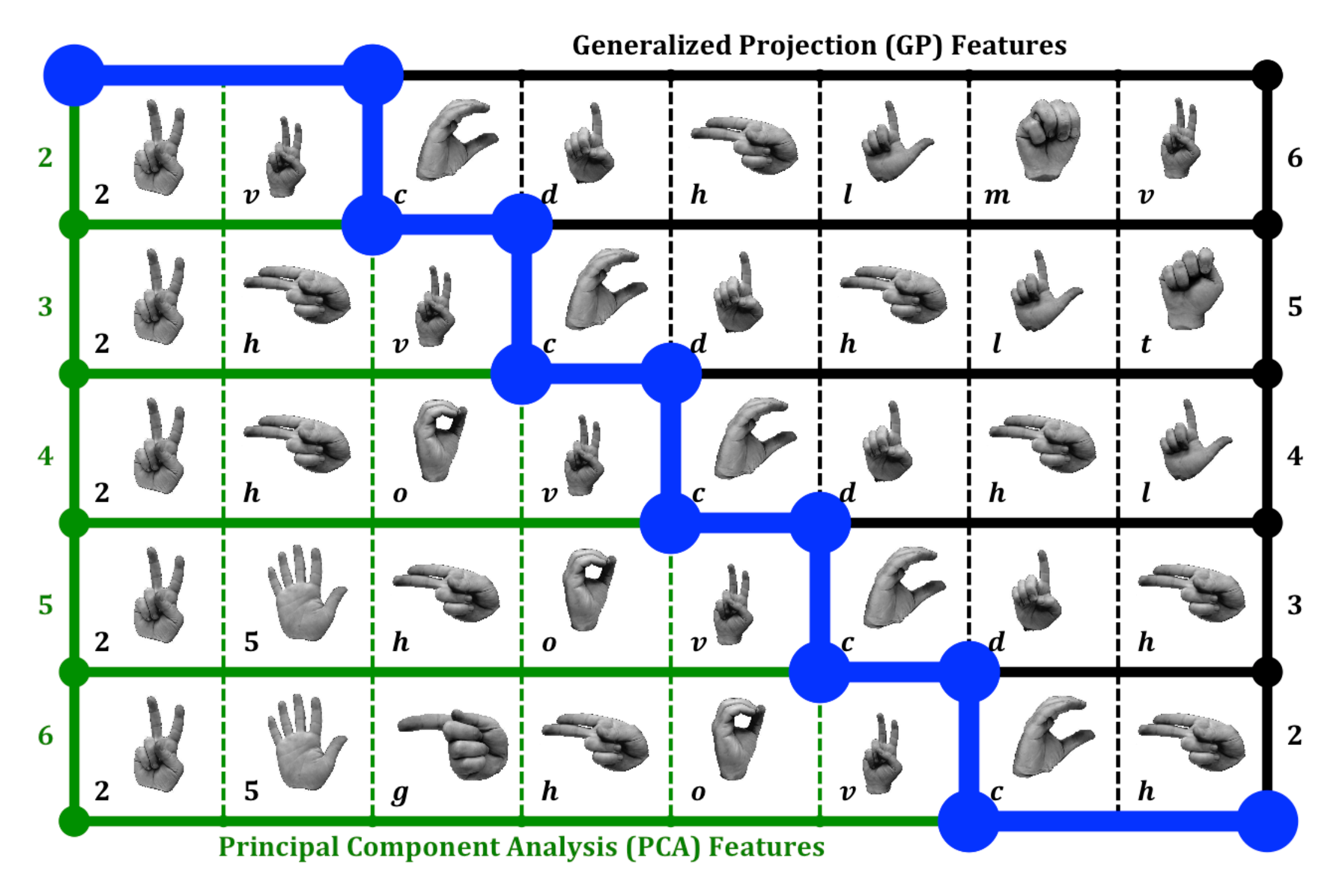}} \\
		\subfloat[][$\alpha = 0.75$]{\la{fig:best75}\includegraphics[width=0.45\linewidth]{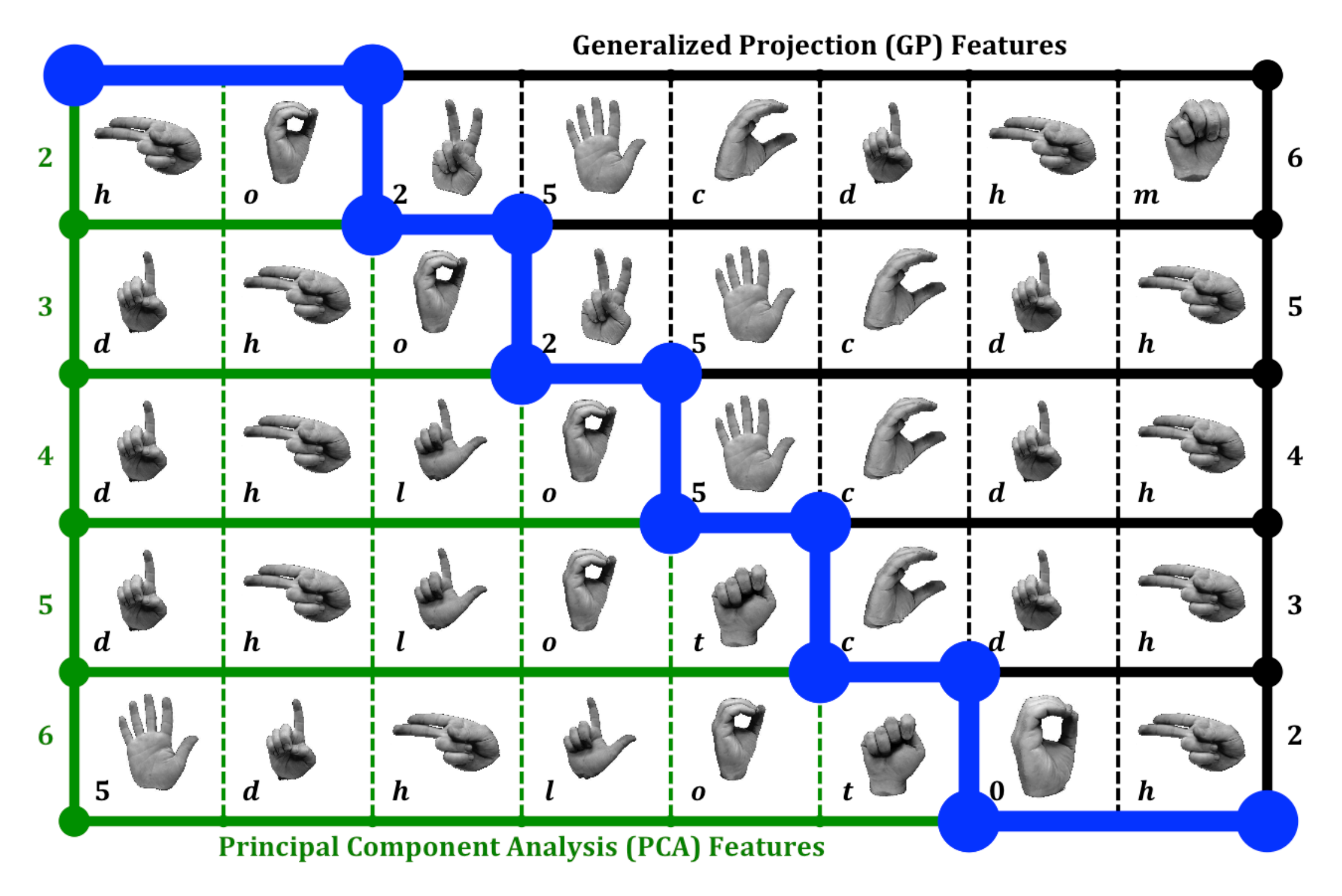}} \quad
		\subfloat[][$\alpha = 1$]{\la{fig:best100}\includegraphics[width=0.45\linewidth]{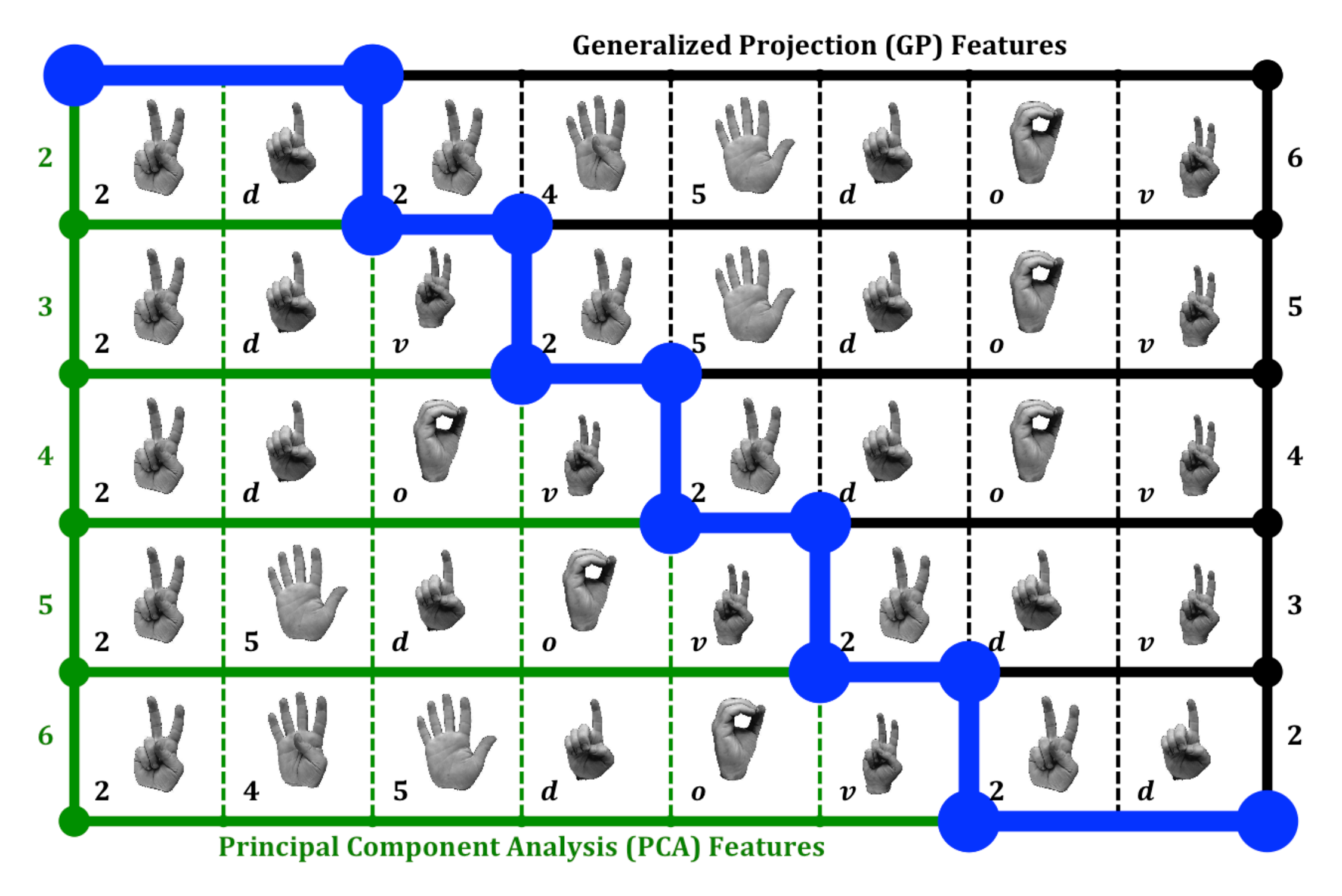}}
	\ec
 	\caption{These are the best lexicons of sizes $n=2-6$ using the PCA (green) and GP (black) feature extraction methods.  
The lexicons of size $n$ are read across each row; the blue line separates the two feature selection methods.  
The relative influence that the subjective considerations for which gestures are better than others is controlled by the weighting factor $\alpha$.  
When $\alpha = 1$, the total measure (TM) comes solely from the subjective measure (SM), and is not swayed by the EDRM.}
	\la{fig:bestsubj}
\end{figure*}

\section{Conclusions and Outlook}

Finding a limited number of best gestures in a large lexicon (library) of gestures for a particular software application is of growing important for consumer electronics and computer vision applications.  To have broad technological appeal, such optimal selection must not only be accomplished through an objective algorithm, but must include consumer preference, or subjective constraints, as well.
The manner in which feature data from the gesture images lie in feature space provides excellent clues as to how well a general recognition algorithm will perform.  
As is expected, well-clustered features of the same class that are also well-separated from the features of other classes are ideal for consistent and  accurate gesture recognition by computer algorithms.  
When the feature data is well-clustered and well-separated, the perils of {\em over-fitting} the data are diminished and the complications of working with high-dimensional feature data is abated.  
However, many features are not well separated, and more sophisticated techniques, like those outlined here, play a key role in providing an objective and subjective measure of the best gesture selection process.

One does not always have control over what features are available for a given recognition problem, but often there are many possible candidates for features.  In such cases, it is best to remove weaker features and select the best features when attempting to find the best gesture set of size $n$.  
The {\em feature selection weakness} (FSW) is one feature strength measure well-suited for determining which features will be well-clustered and well-separated in feature space.  

The {\em ellipsoidal distance ratio metric} (EDRM) is a measure devised to reward optimal feature clusterings in feature space by assigning a value to each gesture class pairing that accounts for their respective separation compared to their individual clustering sizes.  
This measure has been shown to match the true average recognition rate results that decide which combination of $n$ gestures is best, which are obtained by actually completing entire recognition process.  
The EDRM value only needs the feature data to provide a well-educated guess as to what is the best lexicon of size $n$.  

The EDRM is not without some failings.  
It can suffer from the effects of the {\em curse of dimensionality}, since the ellipsoid dimensions are determined by the number of features extracted.  
This means that as the dimensions increase, more data is needed to make accurate predictions about the true recognition rates.  
Sorting through all the possible combinations of $n$ gestures using the EDRM is an {\em NP hard} problem, which can quickly become intractable as the size of the entire lexicon of available gestures increases; though, subjective constraints applied to entire set of all viable gestures often can eliminate this problem.  
Also, the EDRM is designed for filter-based feature selection methods where the features are extracted generally, independent of the classifier.  
Wrapper-based and embedded feature selection methods are directly tied to the specific classifier and so may not be as generalizable and can be computationally expensive.  
Along with this, the EDRM method of determining best gestures does not account for more sophisticated learning algorithms that can distinguish classes separated in complicated, nonlinear ways, which may work very well for specific applications.  

One major advantage to using the EDRM is the ease of incorporating subjective considerations for which gestures are best.  
A simple $[0,1]$ rating scale can be used to rank individual gestures, and the subjective strength of class pairings is calculated by averages, which keeps subjective measure in the  EDRM range.  
These subjective rankings come from ergonomic and vernacular considerations that are mindful of the one who articulates the gestures, and account for ease of articulation, comfort, cultural norms, and suitability to the given application.  


\section*{Acknowledgements}  
J. N. Kutz acknowledges support from the National Science Foundation (NSF)  (DMS-1007621) and the US Air Force Office of Scientific Research (AFOSR) (FA9550-09-0174).

{\small
\bibliographystyle{elsarticle-num}
\bibliography{BestGestbib}
}

\end{document}